\documentclass[10pt,twocolumn,letterpaper]{article}

\usepackage{wacv}
\usepackage{times}
\usepackage{epsfig}
\usepackage{graphicx}
\usepackage{amsmath}
\usepackage{amssymb}
\usepackage[export]{adjustbox}
\usepackage{wrapfig}
\usepackage{xcolor}
\usepackage{graphicx, subcaption}
\usepackage{lipsum}

\def\wacvPaperID{412} 

\wacvfinalcopy 

\ifwacvfinal
\def\assignedStartPage{1} 
\fi

\ifwacvfinal
\else
\fi

\begin{document}

\title{Two-Level Adversarial Visual-Semantic Coupling for Generalized Zero-shot Learning}
\author{Shivam Chandhok\\
Indian Institute of Technology\\
Hyderabad,India\\
\and
Vineeth N Balasubramanian\\
Indian Institute of Technology\\
Hyderabad,India
}

\maketitle

\begin{abstract}
The performance of generative zero-shot methods mainly depends on the quality of generated features and how well the model facilitates knowledge transfer between visual and semantic domains.
The quality of generated features is a direct consequence of the ability of the model to capture the several modes of the underlying data distribution. 
To address these issues, we propose a new two-level joint maximization idea to augment the generative network with an inference network during training which helps our model capture the several modes of the data and generate features that better represent the underlying data distribution. This provides strong cross-modal interaction for effective transfer of knowledge between visual and semantic domains.
Furthermore, existing methods train the zero-shot classifier either on generate synthetic image features or latent embeddings produced by leveraging representation learning. In this work, we unify these paradigms into a single model which in addition to synthesizing image features, also utilizes the representation learning capabilities of the inference network to provide discriminative features for the final zero-shot recognition task. We evaluate our approach on four benchmark datasets i.e. CUB, FLO, AWA1 and AWA2 against several state-of-the-art methods, and show its performance. We also perform ablation studies to analyze and understand our method more carefully for the Generalized Zero-shot Learning task.
\end{abstract}

\vspace{-15pt}
\section{Introduction}
\label{intro}
\vspace{-5pt}
Practical settings require recognition models to have the ability to learn from few labeled samples and be extended to novel classes where data annotation is infeasible or data of new object classes are included with time. However, deep learning models are not suited directly for such settings due to their reliance on labeled data during training. On the other hand, humans perform well under such conditions due to their capability to transfer semantics and recast information from high-level descriptions to visual space, enabling them to recognize objects that they have never seen before. Zero-shot learning (ZSL) aims to bridge this gap by providing recognition models with the capability to classify images of novel classes that have not been seen during the training phase. The model is given access to semantic description of the novel unseen classes during training (such as embeddings of attributes of the classes) and is expected to recognize unseen class images by knowledge transfer between visual and semantic domains.


Based on the classes that a model sees in the test phase, the ZSL problem is generally categorized into two settings: \textit{conventional} and \textit{generalized zero-shot}. In conventional ZSL, the image features to be recognized at test time belong only to unseen classes. In the generalized ZSL (GZSL) setting, the images at test time may belong to both seen or unseen classes. The GSZL setting is practically more useful and challenging when we compare to the conventional setting, since the assumption that images at test time come only from unseen classes need not hold. We aim to address the generalized zero-shot learning problem in this work.


A potential approach to address generalized zero-shot learning is to utilize generative models to generate features for unseen classes and reduce the zero-shot problem to a supervised learning problem \cite{thirteen, fourteen, fifteen, sixteen, eighteen}. Most existing methods in this direction simply use a unidirectional mapping by generating visual features conditioned on semantic attributes. However, it has been shown that such methods that rely on unidirectional mapping lose out a tight visual-semantic coupling which is crucial for zero-shot recognition \cite{gdan,dascn}. To address this issue, more recent approaches such as \cite{gdan,dascn} have proposed to use bidirectional mapping between visual and semantic domains to enhance zero-shot recognition performance.

In this work, we propose a holistic unified approach for bidirectional mapping that provides a tight coupling between the semantic and visual spaces, and is also expressive enough to capture the complex distributions of the underlying data. Our key contributions are as follows (Figure \ref{fig:my_label} summarizes our overall framework):
\begin{figure*}[h]
    \centering
    \includegraphics[scale=0.5]{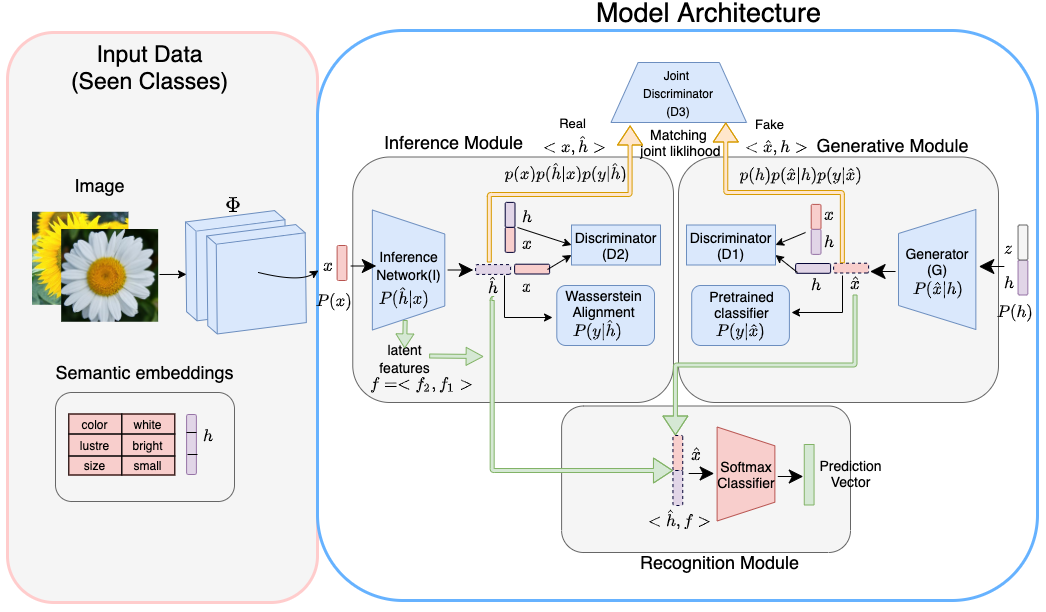}
    \caption{Network architecture for our proposed methodology. The proposed pipeline consists of a Generative module, Inference module, Recognition module and a Joint Discriminator. The model is trained on seen class visual features and semantic attributes. The feature extractor backbone network $\Phi$ is used to extract visual features from images. The vectors generated by our model are shown with dotted outline. The final softmax classifier is trained on synthesized features $\hat x$ and representation from inference network($<\hat h,f>$) as shown.}
    \label{fig:my_label}
\end{figure*}
\noindent (1) Unlike most existing methods that use only a generative module, we augment the generative network with an inference network, and train them together to maximize the joint likelihood of visual and semantic features. Learning the inference network jointly with the generative model helps us capture the underlying modes of the data distribution better \cite{ali}. 

\noindent (2) We provide a two-level adversarial training strategy, where we train both generative and inference modules through respective discriminators. We also use an adversarial joint-maximization loss as an additional supervisory signal to enhance the visual-semantic coupling and facilitate better cross-domain information transfer. This helps our model outperform other dual learning based methods which lack such a mechanism.

\noindent (3) We use a novel Wasserstein semantic alignment loss that helps us model the joint distribution of visual and semantic features better, and ensures that the generated semantic features are distributionally aligned with real semantic features, which in turn helps lower loss in the semantic space. 

\noindent (4) Furthermore, we use the discriminative information in latent layers of the inference network to train our final recognition model. This helps provide the final recognition module with representations from both generative and inference modules, and thus enhances performance when compared to earlier approaches that use only the synthesized image features in the final recognition model.

\noindent (5) We perform detailed experimental studies and analysis on Caltech-UCSD Birds (CUB), Oxford Flowers (FLO) and Animals and Attributes (AWA1 and AWA2) datasets. We demonstrate that the proposed method helps in better visual-semantic coupling, and thus obtains state-of-the art performance, outperforming other methods on both fine-grained as well as coarse-grained datasets.

To the best of our knowledge, this is the first effort to employ a joint maximization step in adversarial training to provide deeper visual-semantic coupling for solving generalized zero-shot learning. In addition, the new idea of using an adversarially learned discriminative representation from the latent layers of inference network, along with the generated features from the generator to train the final zero-shot recognition model, significantly improves GZSL performance. The use of a Wasserstein alignment loss to preserve semantics is also the first of its kind to be used in a generative approach to GZSL.

\vspace{-5pt}
\section{Related Work}
\label{related}
\vspace{-8pt}
As stated in Section \ref{intro}, existing work in ZSL can be broadly divided into work on conventional ZSL and work on generalized ZSL (GZSL). This work focuses on the more challenging GZSL setting, and we focus on presenting related literature in GZSL in this section. 

There has been a recent increase in efforts in the field of zero-shot learning with the aim of boosting GZSL performance.
The methods proposed so far can be broadly categorised into
approaches that learn a projection function based on seen class image features \cite{four,five,seven,sync,nine,six,dem,zskl,dcn,twelve} or generative network based methods which aim to synthesize unseen class features reducing GZSL to a standard supervised problem \cite{fourteen,thirteen,gdan,dascn,se,gzlocd,sgal,sixteen,eighteen,zsml,gmn}. We focus on recent related methods in the rest of this section. The authors in \cite{twelve} first proposed to leverage multi-modal learning by learning a joint embedding of image and textual features for GZSL. They utilized a common representation learning along with cross-domain alignment to map and align visual and semantic features in a common latent space.
To alleviate the bias problem, generative methods for GZSL have been proposed. These methods generally combine adversarial loss and classification loss to generate discriminative features for unseen classes conditioned on semantic attributes. Methods like f-clsWGAN \cite{fourteen}, CVAE \cite{fifteen}, \cite{sixteen} used conditional Generative Adversarial Networks (GANs) or Variational Autoencoders (VAE) for generation of unseen class features. \cite{thirteen} tried combining multi-modal learning with generative GZSL approaches and learned a cross-aligned multi-modal VAE to generate latent features for unseen classes and later trained a softmax classifier on latents from all classes. More recently, \cite{eighteen} proposed to combine the strengths of VAEs and GANs by using the decoder of VAE as a generator. On the other hand, GDAN \cite{gdan} and DASCN \cite{dascn} formulates a dual learning framework that uses bidirectional mapping between visual and semantic spaces, and trained the model with adversarial loss and cyclic consistency. All of these efforts showed the need to enforce stronger coupling between the visual domain (images) and the semantic domain (image attributes provided for seen and unseen classes) in different ways. Our work is closest to GDAN \cite{gdan} and DASCN \cite{dascn} in this regard, and also comes in the category of methods that learn a bidirectional mapping. However, there many differences as described below. Importantly, our approach unifies ideas from existing approaches.

In DASCN \cite{dascn}, in the formulation of dual GAN, the visual to semantic mapping network never sees real image features and only has access to the features generated by the primal generator. As pointed out in \cite{twentytwo}, since the generated image features are practically not as good as actual features, this inhibits the ability of the network to make full use of the dual learning paradigm since the flow of information from visual to semantic domains is partial. 
On the other hand, GDAN \cite{gdan} uses a regressor to implement dual learning which maps the generated features back to the attribute space. As pointed out in \cite{dascn}, minimizing L2-norm between generated semantic embeddings and real semantic attributes is weak and unreliable to preserve high-level semantics when using Euclidean distance. Furthermore, in the objective of GDAN, the only way the generative network and regressor interact with each other is via an L2-norm based cyclic loss which does not provide a strong coupling between the visual and semantic domains. 

In contrast, in our formulation, adversarial learning is introduced in a two-level fashion, where both generative and inference modules are first adversarially learned (see Figure \ref{fig:my_label}). We subsequently then introduce a new adversarial joint maximization loss which specifically aims to maximize the joint probability of visual and semantic features. This is achieved through a joint discriminator, which has a slightly different formulation from a traditional discriminator (as used in GDAN \cite{gdan} and DASCN \cite{dascn}). As pointed in \cite{twentytwo}, learning a regressor by minimizing reconstruction loss performs poorly when compared to learning an inference network jointly. This helps our model generates features that better represent the underlying distribution of unseen classes.
Besides, our design provides our inference network with access to real image features, which facilitates stronger cross-domain coupling with improved representations. Also, our Wasserstein semantic alignment loss enables us to preserve semantics and alleviate semantic loss better than L2 loss.
To show the benefits of our method over GDAN and DASCN, we directly compare with them (as well as many other recent methods) on four different GZSL learning benchmark datasets, and show that our method provides the new state-of-the-art for GZSL.

\vspace{-10pt}
\section{Proposed Methodology}
\label{model}
\vspace{-6pt}
We begin our description of the proposed methodology by defining the problem setting and notations, followed by descriptions of each module of our framework.

\vspace{3pt}
\noindent \textbf{Problem Setting:} Given the dataset $\mathcal{D}= \{D^{Tr}, D^{Ts}\}$,  the aim of generalized zero-shot learning is to correctly classify images from both seen and unseen classes during the test phase. Here, $D^{Tr}=\{ (\textbf{x},  y,  \textbf{h}_y )| \textbf{x} \in \mathcal{X}^s,  y \in \mathcal{Y}^s,  \textbf{h}_y \in \mathcal{A} \}$ is the training set,  where $\textbf{x}$ is an image feature,  $\mathcal{X}^s$ is the feature space of seen classes,  $y$ is the label corresponding to $\textbf{x}$,  $\mathcal{Y}^s$ is the set of labels for seen classes,  $\textbf{h}_y$ is the semantic attribute vector for class $y$. Similarly, $D^{Ts}=\{ (\textbf{x},  y,  \textbf{h}_y )| \textbf{x} \in \mathcal{X}^u,  y \in \mathcal{Y}^u,  \textbf{h}_y \in \mathcal{A} \}$ is the test set,  where  $\mathcal{X}^u$ represents the set of image features from unseen classes,  $\mathcal{Y}^u$ represents the set of labels of unseen classes such that $\mathcal{Y}^u \cap \mathcal{Y}^s = \emptyset$. Note that image features are typically extracted using a feature extractor backbone $\Phi$ as shown in Figure \ref{fig:my_label}. The GZSL task can be formally seen as learning the optimal parameters  of a classifier $f : \textbf{x} \rightarrow \mathcal{Y}^u \cup \mathcal{Y}^s$ where x denote the  image features and $\mathcal{Y}^s$, $\mathcal{Y}^u$ denote the set of labels of seen and unseen classes respectively. 

\vspace{3pt}
\noindent \textbf{Mathematical Framework:} GZSL can be formulated as a problem of modeling joint probability $P(\textbf{x}, \textbf{h}, y)$ where \textbf{x}, \textbf{h} and y are as described above. The joint probability $P(\textbf{x}, \textbf{h}, y)$ can be factored in two ways:
\vspace{-5pt}
\begin{equation}
\label{eq:j1}
\textrm{\textbf{F1}}: p(\textbf{x},\textbf{ h}, y) = p(\textbf{x})p(\textbf{h}|\textbf{x})p(y| \textbf{h})
\vspace{-2pt}
\end{equation}
\begin{equation}
\label{eq:j2}
\textrm{\textbf{F2}}: p(\textbf{x}, \textbf{h}, y) = p(\textbf{h})p(\textbf{x}|\textbf{h})p(y|\textbf{x})
\vspace{-4pt}
\end{equation}
While most existing work consider one of these factorizations in their approach, we use both the factorizations, and model each of them using separate modules, viz. the \textit{generative module} and \textit{inference module} as shown in Figure \ref{fig:my_label}.

For modeling \textbf{F1}, we have the marginal $\Pr( \textbf{x} )$  since we have access to the visual features \textbf{x} of seen classes. The second term $ \Pr( \textbf{h} | \textbf{x} )$, the conditional probability of \textbf{h} given the input \textbf{x}, is modeled by an inference network. Thus, Eqn \ref{eq:j1} can be written as:
\vspace{-4pt}
\begin{align}
\label{eq:j1_supp1}
\Pr( \textbf{x},\textbf{h},y )=\Pr(\textbf{x})\Pr(\hat{\textbf{h}}|\textbf{x} )\Pr(y|\hat{\textbf{h}} )
\vspace{-5pt}
\end{align}
where $\hat{\textbf{h}}$ is the output of the inference module. Now, the term $ \Pr( y | \hat{\textbf{h}} )$ can be factorized as: 
\vspace{-4pt}
\begin{align}
\label{eq:j1_supp2}
\Pr( y|\hat{\textbf{h}} )=\Pr(y|\textbf{h} )\Pr(\textbf{h}|\hat{\textbf{h}} )
\vspace{-5pt}
\end{align}
Here, the factor $\Pr(y|\textbf{h} )$ can be taken as unity, since a known $\textbf{h}$ has a direct map to the label. We attempt to use a Wasserstein alignment to ensures that the $\hat{\textbf{h}}$ is close to the original \textbf{h} in this work. Hence, the joint distribution in factorization $\textbf{F1}$ can now be written as: 
\vspace{-4pt}
\begin{align}
\label{eq:j1_final}
\mathcal{P}_{joint} =\Pr( \textbf{x} )\Pr(\hat{\textbf{h}} | x )\Pr(\textbf{h}|\hat{\textbf{h}})
\vspace{54pt}
\end{align}
This joint distribution is modeled by the \textit{Inference Module} (see Figure \ref{fig:my_label} left), which we discuss in detail in Section \ref{inf}.

For modeling \textbf{F2}, we are provided with the marginal $\Pr( \textbf{h} )$,  since we have access to the attributes \textbf{h} of seen classes. The second term $ \Pr( \textbf{x} | \textbf{h} )$, the conditional probability of \textbf{x} given \textbf{h}, is modeled by a generative network. $\Pr( y | \textbf{x} )$ refers to the extra classification constraint that is present in the loss formulation of our generative network (see Eqn \ref{eq_gen_module_loss_function}).
Thus, Eqn \ref{eq:j2} reduces to:
\vspace{-3pt}
\begin{align}
\label{eq:j2_final}
\mathcal{P}_{joint} =\Pr( \textbf{h} )\Pr( \hat{\textbf{x}} | \textbf{h} )\Pr( y | \hat{\textbf{x}} )
\vspace{-5pt}
\end{align}
where $\hat{\textbf{x}}$ is the output of the generative network. This factorization is modeled by the \textit{Generative Module} (see Figure \ref{fig:my_label} right), which we discuss in detail in Section \ref{gen}.

To provide strong coupling between the generative and inference modules, we also train the generative  and inference networks by maximizing the joint probability and matching Eqns \ref{eq:j1_final} and \ref{eq:j2_final}, which we describe later in this section. To this end, we introduce a joint discriminator $D_3$ (see Figure \ref{fig:my_label} top), whose goal is to match the joint probability by discriminating between a combination of generated and real features i.e $<$\textit{real image feature, generated semantic feature}$>$ and $<$\textit{generated image feature, real semantic feature}$>$ (as shown in Figure \ref{fig:my_label}). This is different from a vanilla discriminator (such as $D_1$,$D_2$ in the figure) which only discriminate between generated and real features.
        


\vspace{-4pt}
\subsection{Generative Module}
\label{gen}
\vspace{-4pt}
Given the training data $D^{Tr}$, the objective of the generative module is to learn a feature generating model conditioned on the semantic attribute vectors, i.e., it should be able to generate discriminative image features that represent the underlying data distribution well. We follow the formulation in \cite{fourteen} for our baseline generative network.
The input to generator $G$ and the discriminator $D_{1}$  are semantic attributes \textbf{h} and image features \textbf{x} of seen classes.
The generator $G$  learns a mapping  $G(\textbf{z}, \textbf{h}): \mathcal{Z} \times \mathcal{A}\rightarrow \mathcal{\hat{X}}$ by taking a random Gaussian noise vector \textbf{z} concatenated with the attribute vector \textbf{h} as input and generating image features $\hat{\textbf{x}}$. We use a Wasserstein GAN \cite{thirty} for this purpose with the loss given by: 
\vspace{-3pt}
\begin{align}
\label{eq:wgan}
\mathcal{L}_{WGAN_1} =& \mathbb E[D_1(\textbf{x}, \textbf{h}(y))] - \mathbb E[D_1(\hat{\textbf{x}}, \textbf{h}(y))] -  \\ 
                   & \lambda \mathbb E[\left(||\nabla_{\tilde{\textbf{x}}} D_1(\tilde{\textbf{x}}, \textbf{h}(y))||_2 - 1\right)^2],  \nonumber
\vspace{-3pt}                 
\end{align}
where $\hat{\textbf{x}}=G(\textbf{z},\textbf{ h}(y))$ is the generated feature,  $\tilde{\textbf{x}} = \alpha \textbf{x} +(1-\alpha)\hat{\textbf{x}}$ with $\alpha \sim U(0, 1)$, and $\lambda$ is a weighting coefficient.  
In order to ensure that the features generated from the network are discriminative, in addition to the adversarial loss, the generated features are required to minimize the classification loss evaluated over a softmax classifier pretrained on seen class features as in \cite{fourteen}: 
\vspace{-3pt} 
\begin{align}
\mathcal{L}_{CLS} = -\mathbb E_{\hat{\textbf{x}}\sim p_{\hat{\textbf{x}}}}[\log P(y| \hat{\textbf{x}}; \theta)]
\vspace{-3pt} 
\end{align}
The final loss for the generative module is then given by: 
\vspace{-4pt}
\begin{align}
\label{eq_gen_module_loss_function}
L_{gen}=\min_G \max_D \mathcal{L}_{WGAN_1} + \beta \mathcal{L}_{CLS}
\vspace{-4pt}
\end{align}
where $\beta$ is a hyperparameter weighting the classifier.

\vspace{-4pt}
\subsection{Inference Module}
\label{inf}
\vspace{-5pt}
It has been shown \cite{ali} that augmenting a generative model with an inference network enhances the ability of the model to capture different modes of the data distribution well and generate samples which better represent the underlying distribution. Since the performance of GZSL depends greatly on the quality of synthesized features, we hypothesize that learning an inference network jointly with the generative model will help our model to express the actual data distribution well and generalize to unseen classes better. 

The inference network and the discriminator $D_2$ together form the Inference Module as shown in Figure \ref{fig:my_label}. The goal of the inference network $I(\textbf{x}): \mathcal{X}\rightarrow \mathcal{\hat{A}}$ is to learn a mapping from image/visual space to semantic space. The module learns a network that maps input image features to corresponding semantic attributes i.e. it attempts to infer the semantic attributes that generated the image in the generation module. The discriminator $D_2: \mathcal{A} \times \mathcal{Z} \rightarrow \mathbb{R}$  outputs a real value. This is also learned through a WGAN, and the loss function is given by:
\vspace{-4pt}
\begin{align}
\label{eq:wgan_}
\mathcal{L}_{WGAN_2} =& \mathbb E[D_2(\textbf{h}(y), \textbf{x})] - \mathbb E[D_2(\hat{\textbf{h}}(y), \textbf{x})] -  \\ 
                   & \lambda \mathbb E[\left(||\nabla_{\tilde{\textbf{h}(y)}} D_2(\tilde{\textbf{h}}(y), \textbf{x})||_2 - 1\right)^2],  \nonumber
\vspace{-3pt}
\end{align}
where $\hat{\textbf{h}}=I(\textbf{x})$ is the generated semantic attribute,  $\tilde{\textbf{h}} = \alpha \textbf{h} +(1-\alpha) \hat{\textbf{h}}$ with $\alpha \sim U(0, 1)$, and $\lambda$ is a weighting coefficient. 

In addition,  we use a Wasserstein metric-based alignment loss at the output of the inference network. 
This aims to ensures that the distribution of class centres of output semantic attributes $\hat{\textbf{h}}$ aligns with the distribution of ground truth $\textbf{h}$,  enabling us to preserve semantic information better. 
The overall loss function for this module is then given by:
\vspace{-3pt}
\begin{align}
\label{eq_inference_loss_function}
L_{inf}=\min_I \max_D \mathcal{L}_{WGAN_2}+ \gamma \mathcal{L}_{wasserstein} 
\vspace{-3pt}
\end{align}
where $\mathcal{L}_{wasserstein}$ is computed using the sinkhorn distance as in \cite{tdsl}, and $\gamma$ is a weighting coefficient. More details of this alignment loss term is provided in the Supplementary Section. 

\vspace{-3pt}
\subsection{Adversarial Joint Maximization}
\label{joint}
\vspace{-3pt}
Existing zero-shot approaches have hitherto not considered the use of a third joint maximization step to improve visual-semantic coupling and cross-domain knowledge transfer. We introduce the use of a joint maximization loss as an additional supervisory signal to enhance visual-semantic interaction. 
We match the joint probability of visual-semantic features using a joint discriminator $D_{3}$ as shown in Figure \ref{fig:my_label}. The joint discriminator is formulated as follows:
\vspace{-3pt}
\begin{multline}
\label{eq_joint_D3}
\mathcal{L}_{joint-discriminator} =\mathbb E[D_3(\textbf{x},\hat{\textbf{h}}(y))] - \mathbb E[D_3(\hat{\textbf{x}},\textbf{h}(y))] -  \\ 
\lambda \mathbb E[\left(||\nabla_{\tilde{x}} D_3( \tilde{\textbf{x}},\tilde{\textbf{h}(y)}),\nabla_{\tilde{\textbf{h}}} D_3( \tilde{\textbf{x}},\tilde{\textbf{h}(y)})||_2 - 1\right)^2]
\vspace{-3pt}
\end{multline}
where $\hat{\textbf{h}}=I(\textbf{x})$, $\hat{x}=G(\textbf{z},\textbf{h})$ are the generated semantic attributes and image features respectively. $\tilde{\textbf{x}} = \alpha \textbf{x} +(1-\alpha)\hat{\textbf{x}}$ and $\tilde{\textbf{h}} = \alpha \hat{\textbf{h}} +(1-\alpha){\textbf{h}}$ with $\alpha \sim U(0, 1)$, as before, and $\lambda$ is a weighting coefficient. Note that the joint discriminator is formulated differently form vanilla discriminators $D_{1},D_{2}$, as mentioned earlier. $D_{3}$ aims to discriminate between the pairs $<\textbf{x}, \hat{\textbf{h}}>$ and $<\hat{\textbf{x}},\textbf{ h}>$ which enables it to match ($p( \textbf{x} )p( \hat{\textbf{h}} | \textbf{x} )$) and ($p( \textbf{h} )p( \hat{\textbf{x}} | \textbf{h} )$) (also shown in Figure \ref{fig:my_label} top).

On the other hand, while $D_{3}$ aims to optimizes Eqn \ref{eq_joint_D3}, the generator and inference networks jointly maximize:
\vspace{-3pt}
\begin{multline}
\label{eq_joint_max}
\mathcal{L}_{joint-max} =\mathbb -(E[D_3(\textbf{x},\hat{\textbf{h}}(y))] - \mathbb E[D_3(\hat{\textbf{x}},\textbf{h}(y))] )
\vspace{-3pt}
\end{multline}
Note that the difference between Eqns \ref{eq_joint_max} and \ref{eq_joint_D3} is only a regularizer term which is added to improve $D3$. In summary, the generative and inference modules are jointly trained to optimize the final objective, given as:
\begin{align}
    \mathcal{L}_{\textrm{total}} &\ = \mathcal{L}_{\textrm{gen}}+\alpha_{1} \mathcal{L}_{\textrm{inf}}+\alpha_{2} \mathcal{L}_{\textrm{joint-max}}
\label{eqn:net_loss}
\end{align}

\vspace{-4pt}
\subsection{Recognition module}
\label{recognition}
\vspace{-5pt}
In previously proposed zero-shot methods, once the generative model is trained, it is used to generated images features for the unseen classes.
However, the synthesized image features by themselves might not be discriminative enough to get best GZSL performance.
In order to address this issue and obtain highly discriminative features for training the classifier in the recognition module, we train the final classifier using the adversarially learned representation in the intermediate layers and the output of the inference network. 

To this end, we combine seen class image features from $D^{Tr}$ and synthesized features for unseen classes from the trained generator into a set, $D$. We then pass the image features in $D$ through the pre-trained inference network and get the output $\hat{\textbf{h}}$, as well as internal intermediate features from the inference network. These are concatenated and used to train the final classifier. For fair comparison with other methods, we use a single layered softmax classifier (as used in most earlier efforts).

At test time, for an input image feature $\textbf{x}_{test}$, we pass it through the inference network and get the internal feature vectors $\textbf{f}$ and output $\hat{\textbf{h}}$. We concatenate these with the test image vector (output of generator module), and pass it to the softmax classifier, given by: 
\vspace{-5pt}
\begin{equation}
f(x)= \arg\max_{y\in\mathcal{\tilde{Y}}}P(y|<\textbf{x}_{test}, \textbf{h}, \textbf{f}>;\Theta')
\vspace{-5pt}
\end{equation}
where $\mathcal{\tilde{Y}}=\mathcal{Y}^s\cup \mathcal{Y}^u$ for GZSL, and $\Theta'$ are the parameters of our overall architecture. 

\begin{table*}[h]
\footnotesize
\centering
 
\vspace{-3pt}
\label{tab:gzsl}
\begin{tabular}{|c|ccc|ccc|ccc|ccc|}
\hline
\textbf{Dataset}     & \multicolumn{3}{c|}{\textbf{CUB}} & \multicolumn{3}{c|}{\textbf{FLO}} &
\multicolumn{3}{c|}{\textbf{AWA1}} &
\multicolumn{3}{c|}{\textbf{AWA2}}  \\ \hline
\textbf{Methods}     & U      & S      & H      & U      & S      & H      & U       & S      & H & U       & S      & H      \\ \hline
\textbf{DEM}(CVPR'17)\cite{dem}	 &19. 6 &57. 9 &29. 2 &- &- &- &32. 8 &\textbf{84. 7} &47. 3 &30. 5 &\textbf{86. 4} &45. 1\\
\textbf{ZSKL}(CVPR'18)\cite{zskl}	  &21. 6 &52. 8 &30. 6 &- &- &- &18. 3 &79. 3 &29. 8 &18. 9 &82. 7 &30. 8\\
\textbf{DCN}(CVPR'18)\cite{dcn}	 &28. 4 &60. 7 &38. 7 &- &- &- &- &- &- &25. 5 &\textbf{84. 2} &39. 1\\
\textbf{ALE}(CVPR'13)\cite{four} &23. 7  &62. 8  &34. 4 &13. 3  &61. 6  &21. 9 &16. 8 &76. 1 &27. 5 &\textbf{81. 8} &14. 0 &23. 9\\
\textbf{DEVISE}(CVPR'13)\cite{five}&23. 8 &53. 0 &32. 8 &9. 9 &44. 2 &16. 2 &13. 4 &68. 7 &22. 4 &74. 7 &17. 1 &27. 8\\
\textbf{ESZSL}(CVPR'15)\cite{seven}	 &12. 6 &63. 8 &21. 0 &11. 4 &56. 8 &19. 0 &6. 6 &75. 6 &12. 1  &77. 8 &5. 9 &11. 0\\
\textbf{SYNC}(CVPR'16)\cite{sync}	 &11. 5 &\textbf{70. 9} &19. 8 &- &- &- &8. 9 &\textbf{87. 3} &16. 2  &\textbf{90. 5} &10. 0 &18. 0\\
\textbf{LATEM}(CVPR'16)\cite{nine}	  &15. 2 &57. 3 &24. 0 &- &- &- &7. 3 &71. 7 &13. 3 &77. 3 &11. 5 &20. 0\\
\textbf{SJE}(CVPR'15)\cite{six}	  &23. 5 &59. 2 &33. 6 &13. 9 &47. 6 &21. 5 &\textbf{74. 6} &11. 3 &19. 6  &73. 9 &8. 0 &14. 4\\ \hline
\textbf{CLSWGAN}(CVPR'18)\cite{fourteen}        & 43. 7*    & 57. 7*   & 49. 7*   & 59. 0    & 73. 8   & \textbf{65. 6}   & -     & -   & -    &57. 9 &61. 4 &59. 6  \\
\textbf{CADA-VAE}(CVPR'19)\cite{thirteen}        &53. 5     & 51. 6   & 52. 4*   & -    & -   & -   & \textbf{72. 8}     & 57. 3   & \textbf{64. 1}  &75. 0 &55. 8 &63. 9      \\
\textbf{VSE}(CVPR'19)\cite{vse}      &39.5* &68.9* &50.2*    & -    & -   & -   & -     & -   & -      &45.6 &88.7 &60.2\\

\textbf{GZLOCD}(CVPR'20)\cite{gzlocd}        &44. 8*     &59. 9*    &51. 3*    & -    & -   & -  & -    & -   & -   & 59.5     & 73.4   & 65.7  \\
\textbf{GDAN}(NIPS'19)\cite{gdan}       & 39. 3*   & 66. 7*   & 49. 5*   & -   & -   & -   & -    & - & -   &32. 1 &67. 5 &43. 5    \\
\textbf{DASCN}(NIPS'19)\cite{dascn}       &45. 9*     &59. 0*    &51. 6*    & -    & -   & -   & 59. 3     & 68. 0   & 63. 4 &- &- &-      \\
\textbf{SGAL}(NIPS'19)\cite{sgal}        & 40. 9*   & 55. 3*   & 47. 0*    & -    & -   & -   & 52. 7     & 75. 7   & 62. 2   &55. 1 &81. 2 &65. 6    \\ 
\textbf{SE-GZSL}(CVPR'18)\cite{se}       & 41. 5    & 53. 3   & 46. 7    & -    & -   & -   & 56. 3     & 67. 8   & 61. 5  &58. 3 &68. 1 &62. 8     \\ 
\textbf{CycWGAN}(ECCV'18)\cite{sixteen}       & 47. 9   & 59. 3   & 53. 0   & \textbf{61. 6}   & 69. 2   & 65. 2   & 59. 6    & 63. 4 & 59. 8  &59. 6 &63. 4 &59. 8     \\
 
 \textbf{f-VAEGAN}(CVPR'19)\cite{eighteen}    &48. 4     & 60. 1   & 53. 6   & 56. 8    & \textbf{74. 9}   & 64. 6   & -     & -   & -   &57. 6 &70. 6 &63. 5     \\
 
 
\textbf{ZSML}(AAAI'20)\cite{zsml} &60. 0 &52. 1 &55. 7    & -    & -   & - &57.4 &71.1 &63.5  &58.9 &74.6 &65.8 \\

 \hline
\textbf{Ours}        &\textbf{61.2} &57.7 &\textbf{59.4}    &\textbf{60. 6}     &\textbf{81. 1}    &\textbf{69. 4}    &60.5 &71.9,  &\textbf{65.7} &59.4, &74.2, &\textbf{66.0}\\
\textbf{Ours-312}        &51.8*     &60. 0*   & \textbf{55. 6*}    &\textbf{60. 6}     &\textbf{81. 1}    &\textbf{69. 4}    &60.5 &71.9,  &\textbf{65.7} &59.4, &74.2, &\textbf{66.0}\\ \hline
\textbf{Ours(using $\Phi_{2}$)}        &\textbf{64.7}     &65.9    & \textbf{65.35}    &-     &-    &-    &- &-,  &- &62.6 &75.6 &\textbf{68.5}\\
\textbf{Ours-312(using $\Phi_{2}$)}        &55.6*     &\textbf{67.50}*   & \textbf{61.0*}    &-     &-    &-    &- &-,  &- &62.6 &75.6 &\textbf{68.5}\\\hline

\end{tabular}
\caption{GZSL performance comparison with several baseline and state-of-the-art methods. For fair comparison, all results
reported here are \textit{without fine-tuning} the backbone ResNet101 feature extractor. We measure Top-1 accuracy on Unseen(U), Seen(S) classes and their Harmonic mean(H). Best results are highlighted in bold. * indicates result on CUB dataset with only 312 dim attributes (included for fair comparison with other work that use this setting)}
\label{fig:results}
\end{table*}
\section{Experiments and Results}
\label{sec_expt_results}
\vspace{-5pt}
In this section,  we conduct extensive experiments on four public benchmark datasets under the generalized zero-shot learning setting.
We compare our model with several baselines and state-of-the-art methods on four benchmark datasets: \textbf{CUB}~\cite{cub}, \textbf{FLO}~\cite{flo}, \textbf{AWA1}~\cite{awa} and \textbf{AWA2}~\cite{awa}.  Among these datasets, AWA1 and AWA2 are coarse-grained, while FLO and CUB are fine-grained datasets. We follow the standard training/validation/testing splits and evaluation protocols, as in \cite{twentysix}. Following the protocol in \cite{twentysix}, we use ResNet101 as the feature extractor backbone network for fair comparison. We denote this backbone by \textbf{$\Phi_{1}$} henceforth, for convenience. Recently, \cite{sabr} proposed to transform the ResNet-101 image features and use  an 1024-dimensional intermediate representation as input features to overcome hubness and preserve semantic relations. To show the generalizability of our approach, we also evaluate our model on features provided by \cite{sabr}. The feature extractor network corresponding to these 1024-dimensional features is denoted as \textbf{$\Phi_{2}$}, henceforth. We use the class attributes/embeddings provided in \cite{twentysix} for each dataset, which represents the \textbf{h}(.) in our  approach (Fig \ref{fig:my_label}). For CUB and FLO datasets,  we use additional 1024-dimensional character-based CNN-RNN features as in \cite{eighteen}\cite{gmn}, unless explicitly stated otherwise. We use the average-per-class top 1 accuracy metric to evaluate our model on a set of classes, as generally followed \cite{twentysix}. In order to evaluate and compare our model in the  GZSL setting, we report the harmonic mean of our model performance on both seen and unseen classes\cite{twentysix}. For fair comparison, we follow the architecture used in \cite{eighteen} for all our components. Due to space constraints, we provide the details of each component in the supplementary material.

\begin{figure*}[h]
    \centering
    \begin{subfigure}[h]{\columnwidth}
        \vskip 0pt
        \centering
        \includegraphics[width=0.4\linewidth]{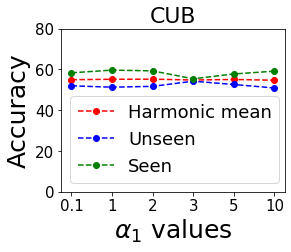}
        \includegraphics[width=0.4\linewidth]{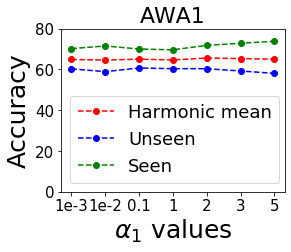}
        \caption{\footnotesize Study of Seen class accuracy, Unseen class accuracy and Harmonic mean by varying $\alpha_{1}$ for fine- and coarse-grained datasets}
        \label{fig:alpha1}
    \end{subfigure}
    \hfill
    \begin{subfigure}[h]{\columnwidth}
        \vskip 0pt
        \centering
        \includegraphics[width=0.4\linewidth]{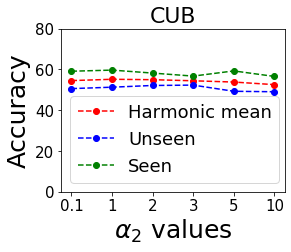}
        \includegraphics[width=0.4\linewidth]{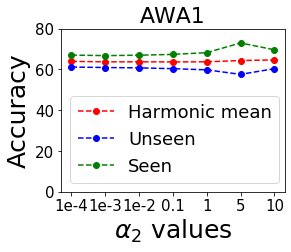}
        \caption{\footnotesize Study of Seen class accuracy, Unseen class accuracy and Harmonic mean by varying $\alpha_{2}$ for fine- and coarse-grained datasets}
        \label{fig:alpha2}
    \end{subfigure}
    \hfill
    \begin{subfigure}[h]{\columnwidth}
        \vskip 0pt
        \centering
        \includegraphics[width=0.4\linewidth]{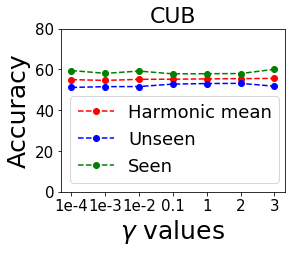}
        \includegraphics[width=0.4\linewidth]{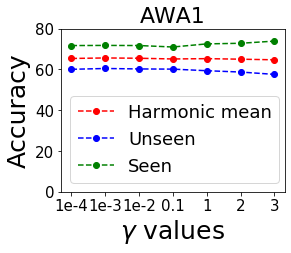}
        \caption{\footnotesize Study of Seen class accuracy, Unseen class accuracy and Harmonic mean by varying $\gamma$ for fine- and coarse-grained datasets}
        \label{fig:gamma}
    \end{subfigure}
    \hfill
    \begin{subfigure}[h]{\columnwidth}
        \vskip 0pt
        \centering
        \includegraphics[width=0.4\linewidth]{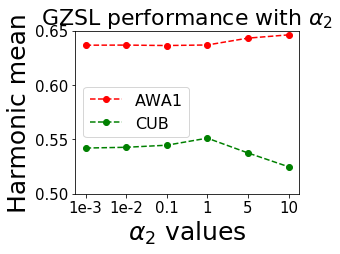}
        \includegraphics[width=0.4\linewidth]{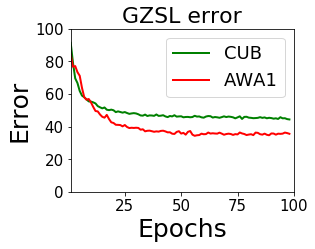}
        \caption{\footnotesize \textit{(Left)} Harmonic mean accuracy for varying $\alpha_{2}$ (plotted on smaller scale for clarity); \textit{(Right)} Training error trajectory over epochs for proposed method}
        \label{fig:gzsl}
    \end{subfigure}
    \hfill
    \begin{subfigure}[h]{\columnwidth}
        \vskip 0pt
        \centering
        \includegraphics[width=0.4\linewidth]{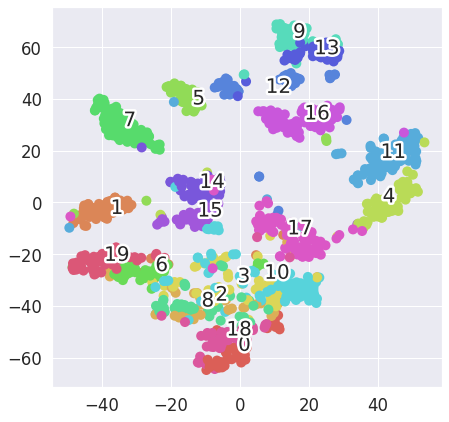}
        \includegraphics[width=0.4\linewidth]{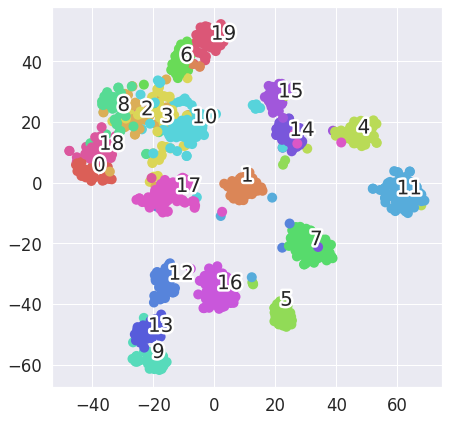}
        \caption{\footnotesize t-SNE visualisations of unseen class features on FLO dataset, which are input to softmax layer of zero-shot classifier (recognition module): \textit{(Left)} f-VAEGAN; \textit{(Right)} Ours}
        \label{fig:cls}
    \end{subfigure}
    \hfill
    \begin{subfigure}[h]{\columnwidth}
        \vskip 0pt
        \centering
        \includegraphics[width=0.4\linewidth]{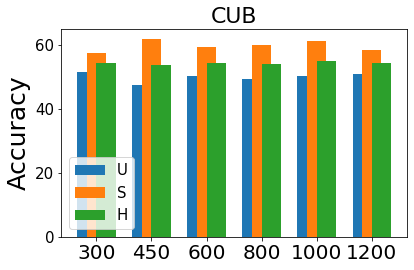}
        \includegraphics[width=0.4\linewidth]{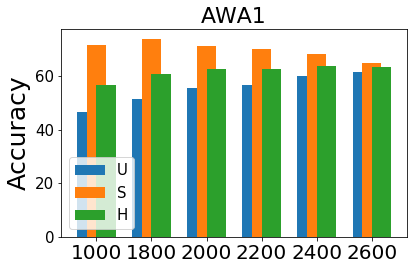}
        \caption{\footnotesize Variation in GZSL performance (S=seen class accuracy; U=unseen class accuracy; H=harmonic mean) with number of synthesised features for unseen classes}
        \label{fig:num}
    \end{subfigure}
    \vspace{-8pt}
    \caption{Ablation Studies and Analysis}
    \vspace{-8pt}
    \label{fig:2}
\end{figure*}

\paragraph{Results:} Table \ref{fig:results} shows the performance comparison of our model with multiple baselines and state-of-the-art methods in GZSL. The table is divided into two sections, which show the performance of non-generative (top) and generative GZSL approaches (bottom) respectively.For fair comparison,all results reported are without fine-tuning the backbone ResNet-101 network.
It can be clearly seen that our methodology consistently outperforms other approaches across all four datasets. Note that GZSL is a challenging problem and most existing methods have not been able to maintain consistently high performance on both fine-grained and coarse-grained datasets. Thanks to the joint-maximization loss and Wasserstein alignment which enables our model to facilitate better cross-domain coupling and learn a useful discriminative representation, our method is able to outperform even other approaches which utilize bidirectional mapping i.e \cite{gdan,dascn} across all datasets with varying granularity.
We also note that our method consistently outperforms approaches like \cite{sixteen,gdan,dascn} which use a cyclic consistency to model visual-semantic interaction.

In Table \ref{fig:results}, we also show the results for our method using $\Phi_{2}$ as the feature extractor backbone for CUB (fine-grained) and AWA2 (coarse-grained) datasets. We see that GZSL performance increases from 59.4$\%$ to 65.35$\%$ in CUB, and 66.0$\%$ to 68.5$\%$ in AWA2. This shows that our model generalizes well to different feature extractor backbones and is not specific to learning from only ResNet-101 features.

\vspace{-5pt}

\begin{figure}[thb]
\centering  
\includegraphics[width=0.54\linewidth]{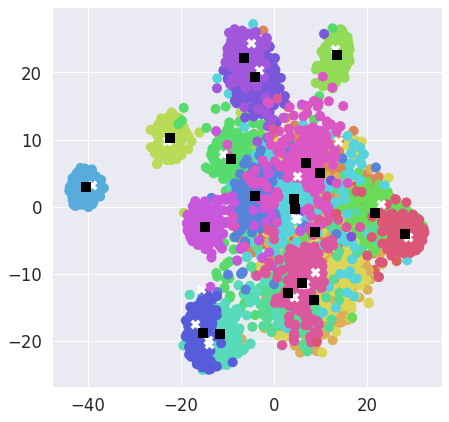}
\vspace{-5pt}
\caption{t-SNE visualization of synthesized image features for unseen classes for FLO dataset. $\blacksquare$ = mean centers of synthesized features; $\times$ (in white) = centres of actual unseen class features \textit{(Best viewed in color, when zoomed in)}}
\vspace{-12pt}
\label{fig:tsne}
\end{figure}

\paragraph{Visualizing Generated Features:} To visualize the unseen class image features generated by our method, we sample 200 synthesized feature vectors for each unseen class for the FLO dataset and plot them using t-SNE as shown in Fig \ref{fig:tsne}. 
In addition, we also show the mean of all synthesized features (represented by $\blacksquare$) and the mean of real unseen classes features (represented by $\times$). It is evident that for most classes, the center/mean of generated synthetic features coincides or is very close to the center of actual unseen class image features. This verifies that our model captures the modes of the underlying distribution well. Furthermore, it can be seen that the features of unseen classes form distinct clusters for most cases which shows the discriminative ability of the generated features.

We note that both at training and test times, our method has time complexity comparable to any other recent GZSL method, and no additional overheads.

\vspace{-5pt}
\section{Ablation Studies and Analysis}
\vspace{-5pt}
We show several ablation studies to show the usefulness of different components as well as the sensitivity of our method to hyperparameter choices in this section. Similar to \cite{thirteen,gdan,sgal,gzlocd}, we use 312-dimensional semantic attributes for our ablation studies on CUB in these studies.

\vspace{3pt}
\noindent \textbf{Relevance of Each Component in our Approach:}
Table \ref{tab_ablation} shows the performance enhancement that each module of our overall architecture brings to GZSL performance. 
\vspace{-4pt}
\begin{itemize}
\setlength\itemsep{-0.5em}
    \item $S1$ corresponds to using only the baseline generative module as explained in Section \ref{gen}. 
    \item $S2$ corresponds to the use of baseline generative module, along with the inference module trained using joint maximization loss. Both $S1$ and $S2$ have the final recognition module trained solely on synthesized image features (i.e $\hat{x}$) without any latent representations from the inference module. 
    \item $S3$ denotes the use of $S2$, along with the use of output of inference network and latent representations from the intermediate layers of the inference module in the recognition module. 
    \item Lastly, $S4$ denotes our complete model, with the Wasserstein alignment loss added to $S3$.
\end{itemize}
\vspace{-8pt}
\begin{table}[h]
\footnotesize
    \centering
    \begin{tabular}{|p{5.6cm}|c|c|}
    \hline \hline
    Model & CUB & AWA1\\
    \hline \hline
    $S1$ = Baseline Generative Module & 51.9 & 61.1   \\
    $S2$ = $S1$ + Inference module + Joint maximization & 52.7 & 62.5  \\     
    $S3$ = $S2$ + Additional features for recognition module  & 54.4 & 65.4  \\
    $S4$ = $S3$ + Wasserstein alignment  &\textbf{55.6} & \textbf{65.7}\\
    \hline \hline
    \end{tabular}
    \vspace{-6pt}
    \caption{Ablation study of different components of our framework on CUB and AWA1. Result reported is harmonic mean accuracy.}
    \vspace{-8pt}
    \label{tab_ablation}
\end{table} 

We draw the following conclusions from Table \ref{tab_ablation}. Training the generative module along with the inference network and joint maximization improves performance for both fine-grained and coarse-grained datasets. The improvement is higher for the coarse-grained dataset, since visual-semantic knowledge transfer becomes more important when classes are farther apart. Utilizing  features from inference network gives a strong boost in GZSL performance for both datasets, showing the importance of good features at final recognition time. 
Lastly, the  Wasserstein alignment also adds improvement for both datasets, although more significantly in fine-grained datasets (CUB). We hypothesize that this is because the classes are close in such datasets, and aligning the distribution of semantic attributes appropriately leads to clearer decision boundary between classes.

\vspace{3pt}
\noindent \textbf{Hyperparameter Choices:}
In Figures \ref{fig:alpha1}, \ref{fig:alpha2}, and \ref{fig:gamma}, we plot the variation in seen class, unseen class and harmonic mean performance with change in hyperparameters $\alpha_{1}$, $\alpha_{2}$, and $\gamma$ respectively - for both coarse-grained and fine-grained datasets. 
In Figure \ref{fig:alpha2}, the best performance is obtained at $\alpha=10$ for the coarse-grained dataset AWA1, and at $\alpha=1$ for the fine-grained dataset CUB. For more careful analysis, the variation of harmonic mean accuracy is also shown in Figure \ref{fig:gzsl}(left) where we plot the GZSL performance on a smaller scale for the sake of clarity. It can be seen that the performance increases with increase in $\alpha_2$ for AWA1 (coarse-grained), while it decreases after a certain point for CUB (fine-grained). This behavior is expected since a higher value of $\alpha_2$ is required in case of coarse-grained datasets, when compared with fine-grained datasets, as it is more difficult to learn from seen classes and generalize to unseen classes for coarse-grained datasets due to classes being more different. In case of Figures \ref{fig:alpha1} and \ref{fig:gamma}, we note that higher values of $\alpha_1$ and $\gamma$ provides the best performance for both CUB and AWA1, showing the importance of the proposed terms. The improvement of higher values of $\alpha_1$ and $\gamma$ is greater in AWA1, which we ascribe to the same reason described above for $\alpha_2$.

\vspace{3pt}
\noindent \textbf{Stability and Generalization:}
Training Generative Adversarial Networks is in general known to be difficult due to an inherently unstable training procedure. In 
Figure \ref{fig:gzsl}(right), we show the training error trajectories over epochs for CUB and AWA1 datasets. We see that the training error smoothly decreases with a stable trend, and reaches convergence within 100 epochs for both fine-grained as well as coarse-grained datasets. 
 
\vspace{3pt} 
\noindent \textbf{Usefulness of Latent Features in Recognition Module:} In order to study the usefulness of using latent feature representations from intermediate layers of the inference network, we plot the representation before the softmax activation layer of the zero-shot classifier (recognition module) of our method and f-VAEGAN-D2, a recent state-of-the-art method, for the FLO dataset in Fig \ref{fig:cls}. We visualize the representations for unseen classes (20 classes), since visualizing the seen classes (82 classes) can be cluttered due to their high number. Notice that the clusters for our method (right subfigure) are more compact than those of f-VAEGAN-D2 (left subfigure) for almost all classes. 
The clusters in f-VAEGAN-D2 show features from one class potentially leaking into other classes, which can result in misclassification. This is however improved in our approach. 

\vspace{3pt}
\noindent \textbf{Variation with Number of Synthesized Features:} Fig \ref{fig:num} shows the performance of our model with varying number of synthesized examples for unseen classes. Note that the trend for harmonic mean is stable with variation in number of  synthesized features for both fine and coarse-grained datasets.

\section{Conclusions}
\label{conclusion}
\vspace{-5pt}
In this work, we propose a unified approach for the generalized zero-shot learning problem that uses a two-level adversarial learning strategy for tight visual-semantic coupling. We use adversarial learning at the level of individual generative and inference modules, as well as use a separate joint maximization constraint across the two modules. In addition, we also show that using the latent representation of intermediate layers of the inference network improves recognition performance. This helps our model unify existing latent representation and generative approaches in a single pipeline. Our contributions in this framework enable us to capture the several modes of the data distribution better and improve GZSL performance by providing stronger visual-semantic coupling. We conduct extensive experiments on four benchmark datasets and demonstrate the value of the proposed method across these fine-grained and coarse-grained datasets. Our future work will include coming up with other ways of performing the joint maximization, as well as considering alignments beyond Wasserstein alignment to improve GZSL performance.

\section{Acknowledgement}
\label{acknowledge}
\vspace{-5pt}
We are grateful to the Department of Science and Technology, India; Ministry of Electronics and Information Technology, India; as well as Intel India for the financial support of this project. We also thank the Japan International Cooperation Agency and IIT-Hyderabad for the provision of GPU servers used for this work.
We thank Joseph KJ and Sai Srinivas for all the insightful discussions, that improved the presentation of this work.

 






{\small
\bibliographystyle{ieee_fullname}
\bibliography{ms}

@article{dem,
author = {Zhang, L. and Xiang, T. and and Gong, S},
title = {Learning a deep embedding model for zero-shot learning },
journal = {CVPR},
 
year = 2017
}

@article{zskl,
author = {Zhang  H. and Koniusz, P.},
title = { Zero-shot kernel learning },
journal = {CVPR},
 
year = 2018
}

@article{dcn,
author = {Liu, S. and Long, M. and Wang; and Jordan, M. I.},
title = {Generalized zero
shot learning with deep calibration network.},
journal = {NIPS},
 
year = 2018
}

@article{seven,
author = {B. Romera-Paredes and P. H. Torr. },
title = {An embarrassingly simple
approach to zero-shot learning.},
journal = {ICML},
year = 2015
}

@article{gzlocd,
author = {Rohit Keshari and Richa Singh
 and Mayank Vatsa},
title = {Generalized Zero-Shot Learning Via Over-Complete Distribution},
journal = {CVPR},
year = 2020
}

@article{six,
author = {Z. Akata and S. Reed and D. Walter and H. Lee and B. Schiele.},
title = {Evaluation of output embeddings for fine-grained image classification.},
journal = {CVPR},
year = 2015
}

@article{five,
author = {A. Frome and G. S. Corrado and J. Shlens and S. Bengio and J. Dean, M. A.
Ranzato and T. Mikolov},
title = {Devise: A deep visual-semantic
embedding model.},
journal = {NIPS},
year = 2013
}

@article{four,
author = {Z. Akata and F. Perronnin and Z. Harchaoui and C. Schmid.},
title = {Label embedding for image classification},
journal = {TPAMI},
year = 2016
}

@article{sync,
author = {Changpinyo, S. and Chao, W.-L. and Gong, B.; and Sha, F.},
title = {Synthesized classifiers for zero-shot learning},
journal = {CVPR},
year = 2016
}

@article{nine,
author = {Y. Xian and Z. Akata and G. Sharma and Q. Nguyen and M. Hein and
B. Schiele},
title = {Latent embeddings for zero-shot classification.},
journal = {CVPR},
year = 2016
}

@article{gdan,
author = {He Huang and Changhu Wang and Philip S Yu and Chang-Dong Wang.},
title = {Generative dual adversarial network for generalized zero-shot learning.},
journal = {CVPR},
year = 2019
}

@article{twelve,
author = {Y.-H. H. Tsai and L.-K. Huang and R. Salakhutdinov.},
title = {Learning
robust visual-semantic embeddings.},
journal = {ICCV},
year = 2017
}

@article{thirteen,
author = {Edgar Schonfeld and Sayna Ebrahimi and Samarth Sinha and Trevor Darrell and Zeynep Akata},
title = {Generalized zero-and few-shot learning via
aligned variational autoencoders},
journal = {CVPR},
year = 2019
}

@article{fourteen,
author = {Yongqin Xian and Tobias Lorenz and Bernt Schiele and and Zeynep Akata.},
title = {Feature generating networks for zero-shot learning},
journal = {CVPR},
year = 2018
}

@article{fifteen,
author = {Ashish Mishra and Shiva Krishna Reddy and Anurag Mittal and Hema A
Murthy.},
title = {A generative model for zero shot learning using conditional
variational autoencoders.},
journal = {CVPRW},
year = 2018
}

@article{sixteen,
author = {Rafael Felix and Vijay BG Kumar and Ian Reid and Gustavo Carneiro.},
title = {Multi-modal cycle-consistent
generalized zero-shot learning.},
journal = {In Proceedings of the European Conference on Computer Vision},
year = 2018
}

@article{dascn,
author = {Jian Ni and Shanghang Zhang and Haiyong Xie},
title = {Dual adversarial semantics-consistent network for generalized zero-shot learning},
journal = {NeurIPS, },
year = 2019
}

@article{eighteen,
author = {Xian, Y. and Sharma, S. and Schiele, B. and Akata, Z.},
title = {A feature generating framework for any-shot learning},
journal = {CVPR },
year = 2019
}

@article{ali,
author = {V. Dumoulin and I. Belghazi and B. Poole and A. Lamb and M. Arjovsky and O. Mastropietro and A. Courville.},
title = {Adversarially
learned inference.},
journal = {ICLR },
year = 2017
}

@article{sabr,
author = {Akanksha Paul and Narayanan C. Krishnan and Prateek Munjal},
title = {Semantically Aligned Bias Reducing Zero Shot Learning},
journal = {CVPR },
year = 2019
}

@article{twenty,
author = {Yunchen Pu and Shuyang Dai and Zhe Gan and Weiyao Wang and Guoyin Wang and Yizhe Zhang and Ricardo
Henao and Lawrence Carin. },
title = {Jointgan: Multi-domain joint distribution learning with generative
adversarial nets. },
journal = {ICML },
year = 2018
}

@article{twentyone,
author = {M.-Y. Liu and O. Tuzel.},
title = {Coupled generative adversarial
networks },
journal = {NIPS },
year = 2016
}

@article{twentytwo,
author = {Jeff Donahue and Philipp Krähenbühl and Trevor Darrell},
title = {Adversarial feature learning},
journal = {ICLR },
year = 2017
}

@article{twentythree,
author = {M. Mirza and S. Osindero.},
title = {Conditional generative adversarial nets.},
journal = {CoRR },
year = 2014
}

@article{twentysix,
author = {Y. Xian and C. H. Lampert and B. Schiele and Z. Akata.},
title = {Zeroshot learning-a comprehensive evaluation of the good, the
bad and the ugly.},
journal = {TPAMI},
year = 2018
}

@article{twentyseven,
author = {I. Goodfellow and J. Pouget-Abadie and M. Mirza and B. Xu and
D. Warde-Farley and S. Ozair and A. Courville and Y. Bengio.},
title = {Generative adversarial nets},
journal = {NIPS},
year = 2014
}

@article{twentyeight,
author = {A. Radford and L. Metz and S. Chintala.},
title = {Unsupervised representation learning with deep convolutional generative adversarial networks},
journal = {ICLR},
year = 2016
}

@article{zsml,
author = {Vinay Kumar Verma and Dhanajit Brahma and Piyush Rai},
title = {Meta-Learning for Generalized Zero-Shot Learning},
journal = {AAAI},
year = 2020
}

@article{twentynine,
author = {X. Chen and Y. Duan and R. Houthooft and J. Schulman and I. Sutskever
and P. Abbeel.},
title = {Infogan: Interpretable representation learning
by information maximizing generative adversarial nets.},
journal = {NIPS},
year = 2016
}

@article{thirty,
author = {M. Arjovsky and S. Chintala and L. Bottou.},
title = {Wasserstein gan.},
journal = {ICML},
year = 2017
}

@article{thirtyone,
author = {I. Gulrajani and F. Ahmed and M. Arjovsky and V. Dumoulin and
A. Courville. },
title = {Improved training of wasserstein gans.},
journal = {NIPS},
year = 2017
}

@article{sinkhorn,
author = {M. Cuturi.},
title = { Sinkhorn distances: Lightspeed computation of optimal transport},
journal = {NIPS},
year = 2013
}

@article{vse,
author = {Zhu; Pengkai and Wang, H. and Saligrama, V.},
title = {Generalized
zero-shot recognition based on visually semantic embedding.},
journal = {CVPR},
year = 2019
}

@article{cub,
author = {P. Welinder and S. Branson and T. Mita and C. Wah and F. Schroff and S. Belongie and P. Perona.},
title = {Caltech-ucsd birds},

year = 2010
}

@article{awa,
author = {C. H. Lampert and H. Nickisch and S. Harmeling.},
title = {Learning to
detect unseen object classes by between-class attribute transfer.},
journal = {CVPR},
year = 2009
}

@article{flo,
author = {M.-E. Nilsback and A. Zisserman.},
title = {Automated flower classification over a large number of classes},
journal = {ICCVGI},
year = 2008
}

@article{gmn,
author = {Mert Bulent and Sariyildiz,Ramazan Gokberk Cinbis},
title = {Gradient Matching Generative Networks for Zero-Shot Learning},
journal = {CVPR},
year = 2016
}

@article{se,
author = {V. Kumar Verma and G. Arora and A. Mishra and P. Rai.},
title = {Generalized zero-shot learning via synthesized examples.},
journal = {CVPR},
year = 2018
}

@article{sgal,
author = {Hyeonwoo Yu and Beomhee Lee},
title = {Zero-shot Learning via Simultaneous Generating and Learning},
journal = {NIPS},
year = 2019
}

@article{tdsl,
author = {Ziyu Wan
and Dongdong Chen
and Yan Li
and Xingguang Yan
and Junge Zhang
and Yizhou Yu
and Jing Liao},
title = {Transductive Zero-Shot Learning with Visual
Structure Constraint},
journal = {NIPS},
year = 2019
}
}

\end{document}


\newrefcontext[labelprefix=A]
\renewcommand{\thefigure}{\Alph{figure}}
\renewcommand{\thetable}{\Alph{table}}


\appendix

\twocolumn[{%
\renewcommand\twocolumn[1][]{#1}%

\maketitle
\begin{center}
    \centering
    \includegraphics[scale=0.5]{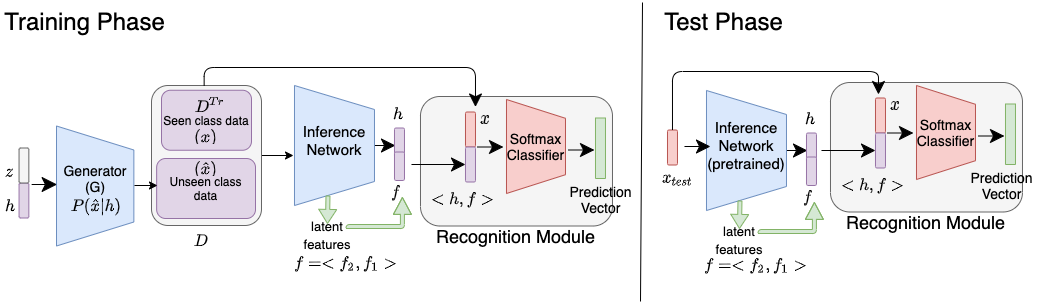}
    \captionof{figure}{Training and testing phase pipelines as explained in Section 3.3 for the proposed framework that uses representation from inference network, along with synthesized features for training recognition module. }
    \label{fig:A}
    
\end{center}%
}]

In this supplementary section, we discuss the following details, which could not be included in the main paper owing to space constraints:
\vspace{-5pt}
\begin{itemize}
\setlength\itemsep{0em}
    \item Implementation details of our experiments (in continuation to Sec 4)
    \item Details describing the Alignment loss (in continuation to Sec 3.2)
    \item Related work on Generative Models (in continuation to Sec 2)
\end{itemize}
We have also added a complete figure, Figure \ref{fig:A}, with both training and testing phases of the recognition module for clarity of understanding.


 





\section{Implementation Details}
\label{intro}
In this section, we describe the implementation details for our methodology, which we could not include in our experiments section. 
The generator ($G$), inference network ($I$) and discriminators ($D_{1},D_{2},D_{3}$) are all implemented using fully connected neural networks. In order to ensure fair comparison, we follow the architecture used in \cite{eighteen} for all our components. Formally, the generator and inference network both consist of 2 dense layers of size 4096 with leaky ReLU activation except at the output layer which has ReLU activation. These layers in the inference network form the latent features $\textbf{f}_{1},\textbf{f}_{2}$ in our recognition module as shown in Figure \ref{fig:A}. The dimension of output layer is 2048 in case of the generator and $d_{h}$ in case of the inference network where $d_{h}$ denotes the dimension of semantic attributes. The three discriminators consist of 2 fully connected hidden layers of size 4096 with leaky ReLU activation.
The noise vector $z$ is sampled from a random unit Gaussian. We find that taking the dimension of noise vector same as that of semantic embeddings works well as in \cite{eighteen}. The generator is updated every 5 discriminator iterations as suggested in WGAN paper ~\cite{thirtyone}.
As for the optimization, we use an Adam optimizer. We use a single layered softmax classifier in our recognition module for fair comparison (most earlier work use this) and simplicity.
Table \ref{tab:my_label} presents the details of hyperparameters used for each of the considered benchmark datasets.

\begin{table}[ht]
\footnotesize
    \centering
    \begin{tabular}{|c|c|c|c|c|c|}
    \hline
    Dataset & $\beta$ &$\lambda$ & $\gamma$ & $\alpha_{1}$ & $\alpha_{2}$\\
    \hline
    \hline
    CUB & 0.01 &10 & 3 & 1 & 2\\
    FLO & 0.01 &10 & 0.01 & 1 & 1\\
    AWA1 & 0.01 &10 & 0.001 & 10 & 2\\
    AWA2 & 0.01 &10 & 0.01 & 5 & 4\\
    \hline
    \end{tabular}
    \caption{Hyperparameters used for different datasets}
    \label{tab:my_label}
\end{table}
\section{Alignment Loss}
In this section, we explain the details regarding of the $L_{wasserstein}$ term in Eqn 11 of the main paper.
We use the sinkhorn distance based lightspeed computation proposed by Cuturi in 2013 \cite{sinkhorn} for computing our alignment loss. The sinkhorn distance metric has also been used for approximating the Wasserstein distance in \cite{tdsl} for a very different projection-based (non-generative) ZSL method for alignment in visual space. 
We instead use the metric to provide distributional alignment in the semantic space which helps us to preserve high-level semantics better and reduce semantic loss. To the best of our knowledge, this has not been done before in ZSL literature. We use the Wasserstein distance to model the joint probability of visual-semantic features better by combining it with adversarial loss in a generative GZSL setting.
Formally, the sinkhorn distance can be written as:
\begin{equation}
    \mathcal{L}_{Wasserstein} = \underset{X}{\min}\sum _ { i , j } \text{dis}_{i j} x_{i j} - \epsilon H(X)
\end{equation}
where $H(X)$ is an entropy-based regularization term and $\text{dis}_ {i j} (\cdot)$ is as defined in \cite{tdsl}. We compute $\text{dis}$ however on an assignment matrix $X$ with entries given by $x_{i j}$, which defines the matching relationship between the class centers of output semantic attributes, $\hat{\textbf{h}}$, and ground truth semantic centers $\textbf{h}$. Here, $i \in A$ and $j \in B$ where $A$ and $B$ are sets of class centres of output semantic attributes and ground truth semantic attributes respectively. This helps compute the term $P(\textbf{h}|\hat{ \textbf{h}})$ in our methodology.
\section{Related Work: Generative Models}
Since ours is a generative approach to GZSL, we briefly present the earlier work in generative models underlying our methodology, for completeness of our discussion on related work. Generative modeling aims to learn the probability distribution of data points such that we can randomly sample data from it. The idea behind Generative Adversarial Networks (GANs) is to learn a generative model to capture an arbitrary data distribution via a min-max training procedure which consists of a generator that synthesizes fake data and a discriminator that distinguishes fake and real data. These models have been used in many interesting computer vision applications especially for image generation \cite{twentyseven,twentyeight,twentynine} and have achieved compelling results. However, GANs are also known for their instability in training, and are known to suffer from the mode collapse problem. In order to mitigate these problems and improve the quality of synthetic samples, methods like WGAN \cite{thirty} and WGAN-GP \cite{thirtyone} have been proposed, which we leverage in this work.
GANs have also been used for tasks like multi-view generation and learning cross-modal representations for downstream tasks like retrieval or semi-supervised classification \cite{twentyeight}
\cite{ali}.
Such generative models can be trained explicitly to model conditional/joint distributions of random variables, which we leverage in this work. For example, \cite{twentythree} shows how such generative models can be used to generate data for a specific class by conditioning them on the label. \cite{ali,twenty,twentyone,twentytwo} show how adversarial training can be used to model joint and utilize the trained model for semi-supervised learning.

\printbibliography